\documentclass[10pt,twocolumn,letterpaper]{article}

\usepackage[]{cvpr} 

\usepackage[accsupp]{axessibility}
\definecolor{cvprblue}{rgb}{0.21,0.49,0.74}
\usepackage[pagebackref,breaklinks,colorlinks,allcolors=cvprblue]{hyperref}

\def\paperID{*****} 
\def\confName{CVPR}
\def\confYear{2025}

\usepackage{xcolor}
\usepackage{graphicx}
\usepackage{amsmath}
\usepackage{amssymb}
\usepackage{algorithm}
\usepackage{algorithmic}
\usepackage{color, colortbl}
\usepackage{multirow,adjustbox}
\newcommand{\ours}{{DySS}\xspace}

\title{\ours: Dynamic Queries and State-Space Learning for Efficient 3D Object Detection from Multi-Camera Videos}

\author{Rajeev Yasarla$\quad$
Shizhong Han$\quad$
Hong Cai $\quad$
Fatih Porikli \\
[2mm]
Qualcomm AI Research\thanks{Qualcomm AI Research is an initiative of Qualcomm Technologies, Inc}\\
{\tt\small \{ryasarla, shizhan, hongcai, fporikli\}@qti.qualcomm.com}
}

\begin{document}
\maketitle
\begin{abstract}
Camera-based 3D object detection in Bird's Eye View (BEV) is one of the most important perception tasks in autonomous driving. Earlier methods rely on dense BEV features, which are costly to construct. More recent works explore sparse query-based detection. However, they still require a large number of queries and can become expensive to run when more video frames are used. 

In this paper, we propose \ours, a novel method that employs state-space learning and dynamic queries. More specifically, \ours leverages a state-space model (SSM) to sequentially process the sampled features over time steps. In order to encourage the model to better capture the underlying motion and correspondence information, we introduce auxiliary tasks of future prediction and masked reconstruction to better train the SSM. The state of the SSM then provides an informative yet efficient summarization of the scene. Based on the state-space learned features, we dynamically update the queries via merge, remove, and split operations, which help maintain a useful, lean set of detection queries throughout the network. 
Our proposed \ours achieves both superior detection performance and efficient inference. Specifically, on the nuScenes test split, \ours achieves 65.31 NDS and 57.4 mAP, outperforming the latest state of the art. On the val split, \ours achieves 56.2 NDS and 46.2 mAP, as well as a real-time inference speed of 33 FPS.
\end{abstract}
\section{Introduction}
\label{sec:intro}

Camera-only 3D detection is key for autonomous driving due to its lower cost as compared to LiDAR and its ability to detect long-range objects. Recent years have seen significant advancements in this area, where the existing state-of-the-art methods can be categorized into two main paradigms.

The first paradigm includes methods that rely on creating dense, Bird’s Eye View (BEV) representation~\cite{li2023bevstereo,philion2020lift,huang2021bevdet,huang2022bevpoolv2,li2022bevformer,yang2023bevformer,park2022time,li2024bevnext}. These methods typically follow a two-stage process: 1) first constructing dense BEV features from multi-camera inputs, and 2) then performing object detection in the BEV space. These methods are computationally intensive, especially when a high resolution is used for the BEV space, and require complex view transformation operations.

\begin{figure}[t!]
\vspace{-5pt}
    \centering
    \includegraphics[width=\linewidth]{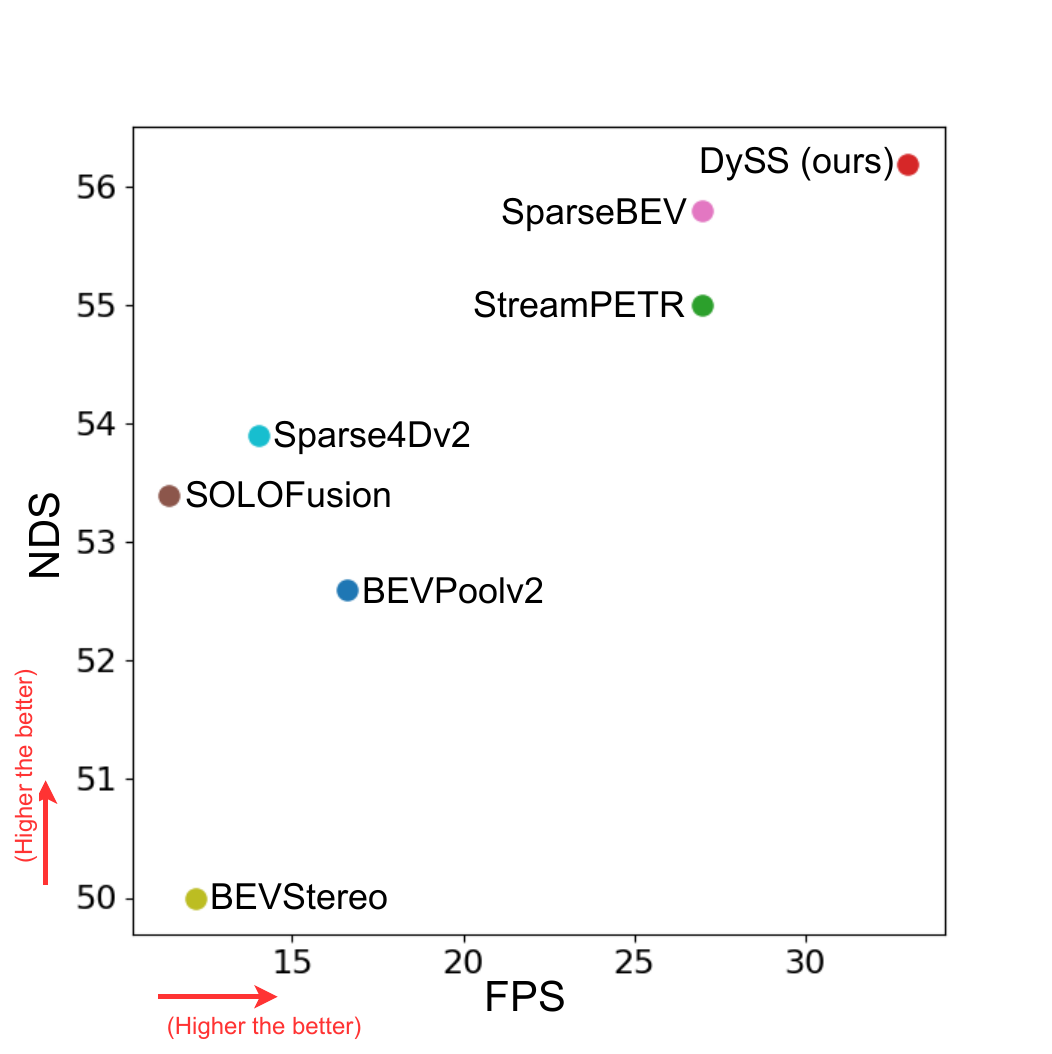} 
    \vskip -8pt 
    \caption{\small \ours~vs. existing SOTA (using ResNet50 backbone) in terms of NDS (nuScenes detection score) and FPS (frames per second, on NVIDIA RTX-3080 GPU), on nuScenes val split. }
    \label{fig:motivation}
    
\end{figure}

Another line of work proposes to use sparse queries initialized in the 3D or BEV space~\cite{carion2020end,wang2022detr3d,liu2022petr,liu2023petrv2,lin2022sparse4d,wang2023exploring,lin2023sparse4dv2,liu2023sparsebev}. By interacting with the image features, these sparse queries learn to capture 3D objects in the scene. 
For instance, DETR3D~\cite{wang2022detr3d} connects these queries to image features through 3D-to-2D projection. 
More recent papers~\cite{wang2023exploring,liu2023petrv2,lin2022sparse4d} further utilize temporal video frames to improve the detection performance. While they are more efficient than dense BEV-based approaches, the existing sparse query-based methods are still computationally expensive as they require sampling several points from each frame for each query and processing these points jointly with attention operations. In particular, the complexity of these methods does not scale well when more video frames are used. For instance, in SparseBEV~\cite{liu2023sparsebev}, for each of the 900 queries, the model samples 4 points from 6 camera views for 8 frames each, and each point contains a 64-dimensional feature. This results in a huge tensor (900$\times$12288) to be processed by attention layers. 


In this paper, we propose a novel approach, \ours, which includes two key strategies to improve both the accuracy and efficiency of 3D object detection, \ie, state-space learning and dynamic queries. First, we leverage state-space modeling to learn the rich spatial-temporal scene information contained in the multi-camera video frames. More specifically, the state-space model (SSM) sequentially consumes the camera frames of consecutive time steps and generates features to summarize the scene. Moreover, we propose to train the SSM with the auxiliary tasks of iterative future prediction and masked reconstruction, which encourage the model to better learn the underlying motion and multi-view correspondence information.   

In addition, we propose to dynamically update, \ie, merge, remove, and split, the queries through the decoder layers based on the learned spatial-temporal features from the SSM. This allows our model to always maintain a relevant and useful set of queries. In contrast, existing works~\cite{wang2023exploring, liu2023sparsebev} use a static set of queries throughout the layers, resulting in redundancy and higher computation costs. 

Our main contributions are summarized as follows.
\begin{itemize}
    \item We propose an efficient and accurate sparse query-based model, \ours, which performs 3D object detection from multi-camera videos in the BEV space.
    \item More specifically, we propose to leverage state-space modeling to efficiently learn the rich spatial-temporal information contained in the multi-camera video frames. We additionally utilize iterative future prediction and masked reconstruction to train the SSM, allowing it to better capture motion and correspondence cues.
    \item Furthermore, we introduce a dynamic query scheme, where the detection queries are updated based on the learned state-space features. This allows \ours to remove the redundant queries and speed up inference.
    \item Our proposed \ours outperforms the latest, existing state-of-the-art methods in terms of both detection accuracy and inference speed. \ours achieves an NDS score of 65.3 on nuScenes~\cite{caesar2020nuscenes} test benchmark and a real-time inference speed of 33 FPS.
\end{itemize}

\begin{figure*}[t!]
    \centering
    \includegraphics[width=0.96\linewidth]{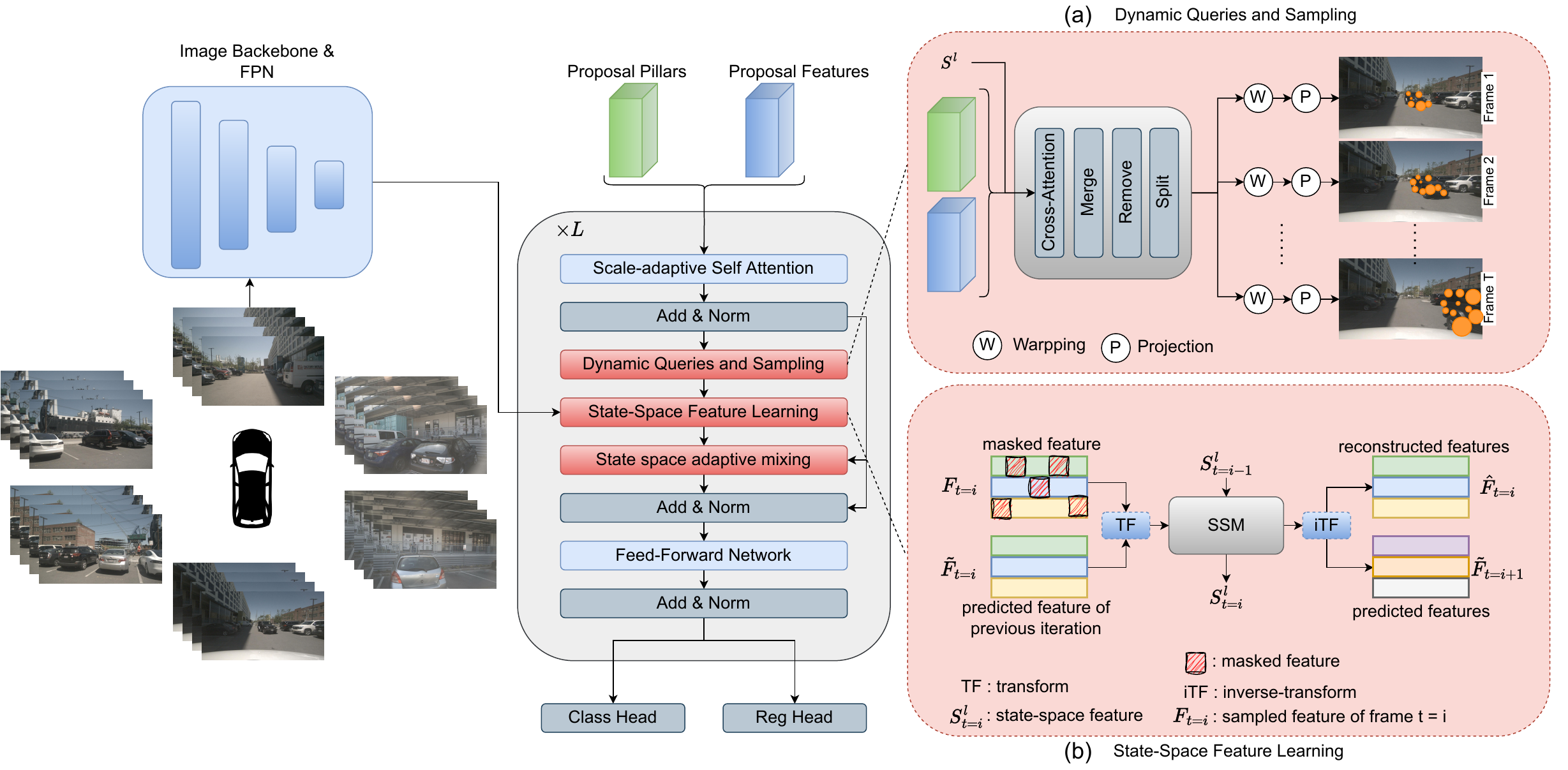} 
    \vskip -5pt 
    \caption{\small Overview of \ours, our proposed efficient 3D object detector for multi-camera videos. With the learnable sparse pillar queries in BEV space, our (a) Dynamic Query Selection and Sampling module removes redundant queries and accelerates inference. The (b) State-Space Feature Learning module enhances the spatial-temporal feature learning through a state-space model. Supervised with future prediction and masked reconstruction, it efficiently learns to capture the motion and scene information.}
    \label{fig:overview}
\end{figure*}
\section{Related Work} 
\label{sec:related_works}
\subsection{Multi-Camera 3D Object Detection in BEV}
Multi-view camera setups are increasingly popular in autonomous driving. Methods for multi-view camera 3D object detection can be divided into two categories: 1) using dense BEV representation and 2)  using sparse representation.

The first group of works convert multi-view perspective images into BEV representation and perform 3D object detection in BEV space~\cite{li2023bevstereo,philion2020lift,huang2021bevdet,huang2022bevpoolv2,li2022bevformer,yang2023bevformer,park2022time,li2024bevnext}. Earlier work like LSS~\cite{philion2020lift} extracts the BEV features by lifting 2D images to 3D space. Based on LSS, BEVDet~\cite{huang2021bevdet} balances accuracy and time-efficiency by optimizing image and BEV resolutions. Its speed is improved via an efficient view transformation by BEVPoolv2~\cite{huang2022bevpoolv2}. BEVFormer~\cite{li2022bevformer} proposed to learn BEV feature representation using spatial-temporal transformers, with further improvement on detection accuracy via perspective 3D detection supervision~\cite{yang2023bevformer}. SOLOFusion~\cite{park2022time} leverages both short-term, high-resolution and long-term, low resolution temporal information to improve depth estimation and detection performance. BEVNeXT~\cite{li2024bevnext} improves BEV feature by incorporating depth estimation modulated with Conditional Random Fields.

The second line of work uses sparse queries to extract features from multi-view images for 3D object detection. Inspired by DETR~\cite{carion2020end}, DETR3D~\cite{wang2022detr3d} learns a set of 3D reference points as object queries to sample the 2D camera features. To simplify this process by removing the complex 2D-to-3D projection and feature sampling, PETR~\cite{liu2022petr} produces 3D position-aware features with 3D position encoding. PETRv2~\cite{liu2023petrv2} and Sparse4D~\cite{lin2022sparse4d} further explore the temporal information from previous frames. StreamPETR~\cite{wang2023exploring} proposes the fusion of object queries in memory, which reduces latency and improves detection accuracy. Sparse4D v2~\cite{lin2023sparse4dv2} further reduces latency and GPU memory by using a recurrent temporal fusion. 
To close the detection accuracy gap with dense BEV methods, SparseBEV~\cite{liu2023sparsebev} proposes a fully sparse 3D object detector by increasing the adaptability of the extracted features.

\subsection{State-Space Models}
State Space Models (SSMs)~\cite{gu2023mamba, gu2021efficiently, smith2022simplified} have emerged as a promising alternative to traditional CNNs and Transformers in deep learning. Initially introduced for efficiently modeling long sequences~\cite{gu2021efficiently}, SSMs have shown significant potential in handling sequential data. They are based on continuous systems that map a 1D sequence, $x(t)\in \mathbb{R}^L$ to an output sequence $y(t)\in \mathbb{R}^L$ through a hidden state $h(t)\in \mathbb{R}^N$. Formally, SSM implements the mapping as: 
\begin{equation}\label{eq:mamba}
    \begin{aligned}
    h(t) = Ah(t-1) + Bx(t)\\
    y(t)=Ch(t)
    \end{aligned}
\end{equation}
where $A\in \mathbb{R}^{N\times N}$ is the evolution matrix of the system, and $B\in \mathbb{R}^{N\times 1},C\in \mathbb{R}^{N\times 1}$ are the projection matrices. In practice, since inputs are discrete, recent methods like Mamba perform discretization, effectively creating a discrete version of the continuous system~\cite{gu2023mamba}. 

In the perception domain, VMamba~\cite{liu2024vmamba} and Vision Mamba~\cite{zhu2024vision} propose to serialize the non-sequential structure of 2D vision data to allow for state-space learning. Voxel Mamba~\cite{zhang2024voxel} and LION~\cite{liu2024lion} leverage SSMs for 3D voxels, which are serialized into a 1D sequence. VideoMamba~\cite{li2024videomamba} utilizes SSM for video understanding by spatial-temporal scanning. Video Mamba Suite~\cite{chen2024video} conducts comprehensive studies on the application of Mamba in video understanding, revealing its strong potential for video perception tasks and promising efficiency-performance trade-offs.

\section{Proposed Approach: \ours} 
\label{sec:method}

Our proposed \ours is a query-based one-stage detector. \ours takes multi-camera videos ($T$ frames per camera) as input. These frames are first processed by an image feature extractor. After that, in the $L$ decoder layers, a set of queries in the BEV space are used to sample locations in the images to extract and interact with relevant features, in order to identify 3D objects. More specifically, in this process, we propose to use a dynamic set of queries which are updated after every decoder layer (Section~\ref{sec:dynamic query}), instead of using a static set of queries as in previous methods~\cite{wang2023exploring, liu2023sparsebev}. In addition, we leverage state-space modeling to more efficiently learn the spatial-temporal information contained in the multi-camera videos (Section~\ref{sec:ssl}), and we utilize the learned features to efficiently enhance the queries via cross-attention (Section~\ref{sec:ss mixing}). Fig.~\ref{fig:overview} provides an overview of our proposed \ours model.


\begin{figure*}[t!]
    \vspace{-0pt}
    \centering
    \includegraphics[width=0.9\linewidth]{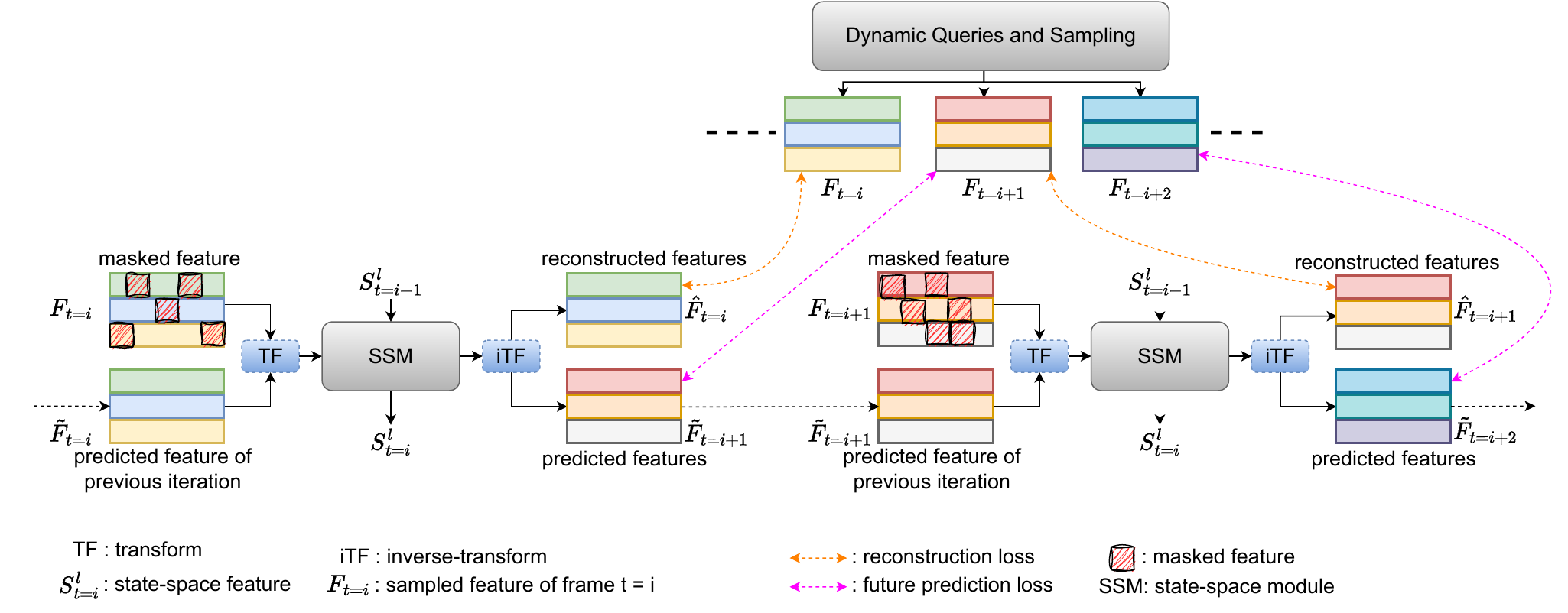} 
    \vskip -8pt 
    \caption{\small Overview of our proposed state-space feature learning module. The state-space model (SSM) sequentially processes the sampled features from each time step. During training, the SSM is supervised with future prediction and masked reconstruction to better learn motion and correspondence information. The losses (bidirectional arrows in figure) are calculated between the predicted features and the output features from Dynamic Queries and Sampling module at all timestamps.}
    \label{fig:state-space}
    \vspace{-0pt}
\end{figure*}

\subsection{Dynamic Query Selection}\label{sec:dynamic query}


Similar to SparseBEV~\cite{liu2023sparsebev}, we first define a set of learnable queries, each of which is represented by its translation $\{x,y,z\}$, dimension $\{w,l,h\}$, rotation $\theta$, and velocity $\{v_x,v_y\}$. These queries are initialized as pillars in the BEV space with $z$ set to 0 and $h$ set to $\sim\!4$m. The initial velocity is set to $\{v_x,v_y\} = \{0,0\}$ and the other parameters $\{x,y,w,i,\theta\}$ are drawn from random Gaussian distributions. Following Sparse R-CNN~\cite{sun2021sparse}, we associate a $D$-dimensional feature with each query to encode the instance information.

Unlike existing SOTA sparse query-based methods~\cite{liu2023sparsebev, wang2023exploring} which all use a fixed set of queries, our proposed \ours leverages dynamic queries. More specifically, at the end of each decoder layer, \ours first performs a cross-attention between the query features and the current state-space features, which provides useful spatial-temporal information to the query features. Then, \ours performs merge, remove, and split operations on the queries sequentially to update them before proceeding to the next decoder layer as shown in Fig.~\ref{fig:overview} (a). Next, we describe these operations in more detail.

\subsubsection{Merge}
After cross-attention with the state-space features, the resulting query features are of $N_q\times D$, where $N_q$ is the number of queries and $D$ is the query feature dimension in the current decoder layer. We compute the covariance matrix based on the query features, $C_{q} \in \mathbb{R}^{N_q \times N_q}$, which captures the similarity among the queries. The covariance matrix and query features are then processed by a linear layer to generate a merge label for each query and an index indicating which queries it should be merged with.


\subsubsection{Remove}
After merging, we remove redundant queries. The query features that have been cross-attended with state-space features are passed to two linear layers to generate a remove label value for each query, ranging from 0 to 1, where 1 indicates the query should be removed. 
Additionally, another linear layer takes the covariance matrix as input and generates a ratio between 0.2 and 0.3, based on which we remove a maximum of 30\% and a minimum of 20\% of the $N_q$ queries.

\subsubsection{Split}
Finally, the remaining queries are processed by the split operator. Similar to the remove operator, two linear layers consume the query features and generates a split label value for each query, ranging from 0 to 1, where 1 indicates the query should be split. Based on the covariance, another linear layer generates a number between 0 and 5, and we split a maximum of 5\% of the $N_{q}$ queries accordingly. When a query is split, we duplicate it.

After the queries are updated using our proposed merge, remove, and split operations, we use them to sample features from the images. Similar to SparseBEV~\cite{liu2023sparsebev}, a linear layer generates a set of sampling offsets $\{\Delta x_i, \Delta y_i, \Delta z_i\}$, which are transformed into 3D sampling points based on the query pillar locations (see Eq.~(4) in~\cite{liu2023sparsebev}). Note that our query features have been enhanced with state-space features, which provide rich spatial-temporal information of the driving scene and enable more accurate sampling. The sampling procedure is similar to that in~\cite{liu2023sparsebev}. The sampled features are then fed into our proposed state-space learning module, which we discuss next.



\begin{figure}[t!]
    \vspace{-0pt}
    \centering
    \includegraphics[width=\linewidth]{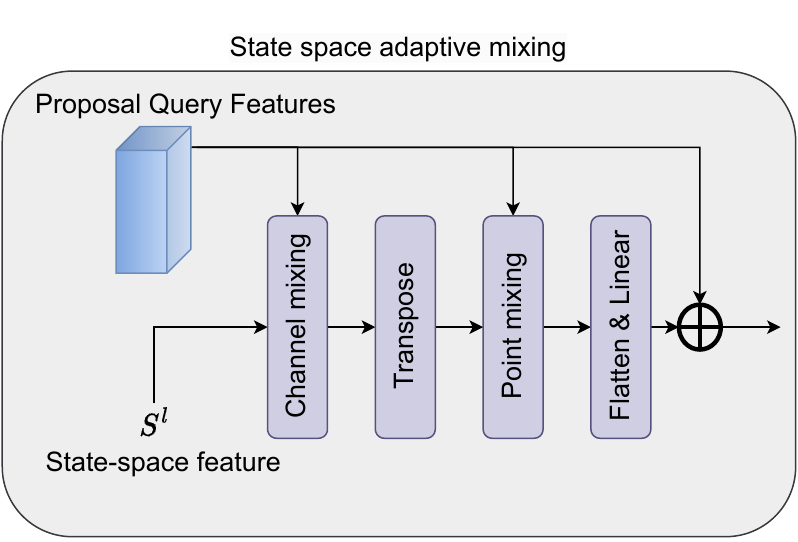} 
    \vskip -8pt 
    \caption{\small Overview of the state-space adaptive mixing module. This module adaptively mixes the proposed query features and state-space features that encode spatial-temporal information, using learned parameters for both channel and point mixing.}
    \label{fig:mixing}
    \vspace{-0pt}
\end{figure}

\subsection{State-Space Feature Learning}\label{sec:ssl}
We employ a state-space model (SSM), such as Mamba~\cite{mamba}, to process the sample features from the multi-camera video frames. Given the sets of sampled features from multiple camera images of consecutive time steps, we sequentially pass them to the SSM, based on which the SSM updates its internal state, $S^l_t=i$, where $i$ is the time step and $l$ is the decoder layer index. 
Specifically, at each time step $t=i$, the SSM consumes not only the sampled features at this time $F_{t=i}$, but also the predicted version of it based on the previous features $\tilde{F}_{t=i}$. It outputs an enhanced version of the sampled features $\hat{F}_{t=1}$ as well as the predicted features for the next time step $\tilde{F}_{t=i+1}$ as shown in Fig.~\ref{fig:overview} (b).

During training, in addition to the standard 3D object detection losses, we further introduce supervision on masked reconstruction and future prediction. Different from inference time where the actual sampled features are passed as part of the input, we provide a masked version of it. The outputs of the SSM are supervised by the ground-truth features at the current time step as well as those from the next time step. Note that conventionally, SSMs are trained in an auto-regressive, step by step way, where the actual observation at each step is used as input. In contrast, we do not rely on such teacher forcing~\cite{alias2017z,spencer2021feedback}, and leverage masked reconstruction and future prediction to enable the model to better learn motion and correspondence information as shown in Fig.~\ref{fig:state-space}; the effectiveness of such auxiliary supervisions has been demonstrated in other video perception tasks, \eg, depth estimation~\cite{yasarla2024futuredepth}.


More formally, the SSM's outputs at each $t=i$, including the enhanced features $\hat{F}_{t=i}$ and predicted features $\tilde{F}_{t+1}$, are supervised as follows:
\begin{equation}\label{eq:mamba loss}
    \begin{aligned}
        \mathcal{L}_{r} &= \frac{1}{T} \sum_{i=1}^T || \hat{F}_{t=i} - {F}_{t=i} ||^2 \\
        \mathcal{L}_{f} &= \frac{1}{T} \sum_{i=1}^T || \tilde{F}_{t=i+1} - {F}_{t=i+1} ||^2,\\
    \end{aligned}
\end{equation}
where $\mathcal{L}_r$ and $\mathcal{L}_f$ are the losses on masked feature reconstruction and future feature prediction, and $T$ is the number of time steps.


Moreover, we propose to optionally perform transform on the features before passing them to the SSM, \eg, Fast Fourier Transform (FFT), discrete cosine transform (DCT), followed by their respective inverse transforms on the resulting output. Such transformation can further improve the learning, as we shall see in the experiments.


\subsection{Adaptive Mixing with State-Space Features}\label{sec:ss mixing}
Inspired by recent adaptive mixing methods~\cite{gao2022adamixer,rao2022amixer}, we perform simple yet effective mixing between the query features and the state-space feature $S^l_{t=T}$, as shown in Fig.~\ref{fig:mixing}.

First, we perform channel mixing.
 \begin{equation}
    \begin{aligned}
        W_c &= \text{Linear}(Q) \in \mathbb{R}^{D \times D}\\
        M_c &= \text{ReLU}(\text{LayerNorm}(S^l_{t=T}W_c)),\\
    \end{aligned}
\end{equation}

Next, we perform point mixing.
 \begin{equation}
    \begin{aligned}
        W_p &= \text{Linear}(Q) \in \mathbb{R}^{P \times P}\\
        M_p &= \text{ReLU}(\text{LayerNorm}(\text{Transpose}(M_c)W_p)).\\
    \end{aligned}
\end{equation}
where $Q$ denotes the query features. $D$ and $P$ are channel dimension and points number. $W_c$ and $W_p$ denote the weights that are adaptively learned from training data.
 
Then, the spatial-temporally mixed features are flattened and combined using a linear layer, with a subsequent residual addition from the query features.
\begin{table*}[t!]
    \centering
    \vspace{-0pt}
    \caption{Comparison with SOTA on nuScenes val. \textdagger~utilizes nuImages~\cite{caesar2020nuscenes} pretraining. }
    \label{tab:val_nuscenes}
    \resizebox{1\linewidth}{!}{
    \begin{tabular}{l|l|c|c|cc|ccccc}
    \hline
    Method & Backbone & Input Size & Epochs & NDS & mAP & mATE & mASE & mAOE & mAVE & MAAE \\ \hline
    PETRv2\cite{liu2023petrv2}  & ResNet50 & $704 \times 256$ & 60 & 45.6 & 34.9 & 0.700 & 0.275 & 0.580 & 0.437 & 0.187 \\ 
    BEVStereo\cite{li2023bevstereo} & ResNet50 & $704 \times 256$ & 90 & 50.0 & 37.2 & 0.598 & 0.270 & 0.438 & 0.367 & 0.190 \\ 
    BEVPoolv2\cite{huang2022bevpoolv2}  & ResNet50 & $704 \times 256$ & 90 & 52.6 & 40.6 & 0.572 & 0.275 & 0.463 & 0.275 & 0.188 \\ 
    SOLOFusion\cite{park2022time}  & ResNet50 & $704 \times 256$ & 90 & 53.4 & 42.7 & 0.567 & 0.274 & 0.511 & 0.252 & 0.181 \\ 
    Sparse4Dv2\cite{lin2023sparse4dv2}  & ResNet50 & $704 \times 256$ & 100 & 53.9 & 43.9 & 0.598 & 0.270 & 0.475 & 0.282 & 0.179 \\ 
    VCD-A\cite{huang2024leveraging} & ResNet50 & $704 \times 256$ & 24 & 56.6 & 44.6 & 0.497 & 0.260 & 0.350 & 0.257 & 0.203\\
    StreamPETR\cite{wang2023exploring}\textdagger & ResNet50 & $704 \times 256$ & 60 & 55.0 & 45.0 & 0.613 & 0.267 & 0.413 & 0.265 & 0.196 \\  
    SparseBEV\cite{liu2023sparsebev}\textdagger& ResNet50 & $704 \times 256$ & 36 & 55.8 & 44.8 & 0.581 & 0.271 & 0.373 & 0.247 & 0.190 \\ 
    \rowcolor{lightgray}\ours (ours)\textdagger& ResNet50 & $704 \times 256$ & 24 & \textbf{56.2} & \textbf{46.2} & 0.537 & 0.268 & 0.338 & 0.239 & 0.177 \\ \hline
    DETR3D\cite{wang2022detr3d}\textdagger  & ResNet101-DCN & $1600 \times 900$ & 24 & 43.4 & 34.9 & 0.716 & 0.268 & 0.379 & 0.842 & 0.200 \\ 
    BEVFormer\cite{li2022bevformer}\textdagger  & ResNet101-DCN & $1600 \times 900$ & 24 & 51.7 & 41.6 & 0.673 & 0.274 & 0.372 & 0.394 & 0.198 \\
    BEVDepth\cite{li2023bevdepth}  & ResNet101 & $1408 \times 512$ & 90 & 53.5 & 41.2 & 0.565 & 0.266 & 0.358 & 0.331 & 0.190 \\ 
    Sparse4D\cite{lin2022sparse4d}\textdagger & ResNet101-DCN & $1600 \times 900$ & 48 & 55.0 & 44.4 & 0.603 & 0.276 & 0.360 & 0.309 & 0.178 \\ 
    SOLOFusion\cite{park2022time}   & ResNet101 & $1408 \times 512$ & 90 & 58.2 & 48.3 & 0.503 & 0.264 & 0.381 & 0.246 & 0.207 \\ 
    HoP-BEVFormer\cite{zong2023temporal}\textdagger  & ResNet101-DCN & $1600 \times 900$ & 24 & 55.8 & 45.4 & 0.565 & 0.265 & 0.327 & 0.337 &  0.194 \\
    StreamPETR\cite{wang2023exploring}\textdagger & ResNet101 & $1408 \times 512$ & 60 & 59.2 & 50.4 & 0.569 & 0.262 & 0.315 & 0.257 & 0.199 \\
    SparseBEV\cite{liu2023sparsebev}\textdagger & ResNet101 & $1408 \times 512$ & 24 & 59.2 & 50.1 & 0.562 & 0.265 & 0.321 & 0.243 & 0.195 \\ 
    \rowcolor{lightgray} \ours (ours)\textdagger& ResNet101 & $1408 \times 512$ & 24 & \textbf{60.4} & \textbf{51.5} & 0.511 & 0.255 & 0.300 & 0.233 & 0.172 \\ \hline
    \end{tabular}
    }
     \vspace{-0pt}
\end{table*}

\section{Experiments}
\label{sec:experiments}

Extensive experiments are conducted to evaluate the proposed \ours approach on standard 3D object detection benchmark and compare it with the existing state of the art (SOTA). Additionally, ablation studies are perform on various designs in \ours.

\subsection{Implementation}
Building upon previous work~\cite{liu2023sparsebev}, we employ common image backbones like ResNet~\cite{he2016deep} and V2-99~\cite{lee2020centermask}. Our decoder contains 6 layers with shared weights. By default, \ours uses 8 frames from each camera with a 0.5-second interval for each inference.

Mamba architecture introduced by \cite{mamba} is used for the proposed state-space learning. To capture both temporal and frequency-domain information, we use two Mamba blocks: one with identity as the transform function to learn time-domain state-space features, and another one using Fast Fourier Transform (FFT) to learn frequency-domain state-space features. For all our experiments, we set the state expansion factor of the SSM to 128.

For training, other than our introduced auxiliary masked reconstruction and future prediction supervisions (Eq.~(\ref{eq:mamba loss})), we follow established 3D detection setting (see~\cite{liu2023sparsebev} for more details). We utilize the Hungarian algorithm to assign ground-truth objects to predictions. Focal loss is used for classification, L1 loss is used for 3D bounding box regression. We use the AdamW optimizer, a global batch size of 8, and an initial learning rate of $2 \times 10^{-4}$ which is adjusted using a cosine annealing policy. 

\begin{table*}[t!]
    \centering
    \caption{Comparison with SOTA on nuScenes test. \textdaggerdbl denotes methods using CBGS\cite{zhu2019class}, where 1 epoch is equivalent 4.5 epochs.}
    \vspace{-5pt}
    \label{tab:test_nuscenes}
    \resizebox{\linewidth}{!}{
    \begin{tabular}{l|l|c|cc|ccccc}
    \hline
    Method & Backbone & Epochs & NDS & mAP & mATE & mASE & mAOE & mAVE & MAAE \\ \hline
    DETR3D\cite{wang2022detr3d}  & V2-99 & 24 & 47.9 & 41.2 & 0.641 & 0.255 & 0.394 & 0.845 & 0.133 \\ 
    PETR\cite{liu2022petr}  & V2-99 & 24 & 50.4 & 44.1 & 0.595 & 0.249 & 0.383 & 0.808 & 0.132 \\ 
    UVTR\cite{li2022unifying}  & V2-99 & 24 & 55.1 & 47.2 & 0.577 & 0.253 & 0.391 & 0.508 & 0.123 \\ 
    BEVFormer\cite{li2022bevformer}  & V2-99 & 24 & 56.9 & 48.1 & 0.582 & 0.256 & 0.375 & 0.378 & 0.126 \\ 
    BEVDet4D\cite{huang2022bevdet4d}  & Swin-B & 90\textdaggerdbl  & 56.9 & 45.1 & 0.511 & 0.241 & 0.386 & 0.301 & 0.121 \\ 
    PolarFormer\cite{jiang2023polarformer}  & V2-99 & 24 & 57.2 & 49.3 & 0.556 & 0.256 & 0.364 & 0.440 & 0.127 \\ 
    PETRv2\cite{liu2023petrv2} & V2-99 & 24 & 59.1 & 50.8 & 0.543 & 0.241 & 0.360 & 0.367 & 0.118 \\ 
    Sparse4D\cite{lin2022sparse4d}  & V2-99 & 48 & 59.5 & 51.1 & 0.533 & 0.263 & 0.369 & 0.317 & 0.124 \\ 
    BEVDepth\cite{li2023bevdepth}  & V2-99 & 90\textdaggerdbl & 60.0 & 50.3 & 0.445 & 0.245 & 0.378 & 0.320 & 0.126 \\ 
    BEVStereo\cite{li2023bevstereo}  & V2-99 & 90\textdaggerdbl & 61.0 & 52.5 & 0.431 & 0.246 & 0.358 & 0.357 & 0.138 \\ 
    SOLOFusion\cite{park2022time}  & ConvNeXt-B  & 90\textdaggerdbl & 61.9 & 54.0 & 0.453 & 0.257 & 0.376 & 0.276 & 0.148 \\ 
    VCD-A\cite{huang2024leveraging}  & ConvNeXt-B & 24 &  63.1 & 54.8 & 0.436 & 0.244 & 0.343 & 0.290 & 0.120 \\
    HoP-BEVFormer\cite{zong2023temporal}  & V2-99 & 24 &  60.3 & 51.7 & 0.501 & 0.245 & 0.346 & 0.362 & 0.105 \\
    StreamPETR\cite{wang2023exploring} & V2-99 & 24 & 63.6 & 55.0 &  0.479 & 0.241 & 0.258 & 0.236 & 0.134 \\ 
    SparseBEV\cite{liu2023sparsebev} & V2-99 & 24 & 62.7 & 54.3 & 0.502 & 0.244 & 0.324 & 0.251 & 0.126 \\ 
    RayDN\cite{liu2024ray} & V2-99 & 60 &  64.5 & 56.5 & 0.461 & 0.241 &  0.322 & 0.239  & 0.114\\
    \rowcolor{lightgray} \ours (ours) & V2-99  & 24 & \textbf{65.3} & \textbf{57.4} & 0.434 & 0.221 & 0.308 & 0.237 & 0.110 \\ \hline
    \end{tabular}
    }
    \vspace{-0pt}
\end{table*}

\begin{table*}[t!]
    \caption{\small Ablation study on nuScenes val with SparseBEV~\cite{liu2023sparsebev} as the baseline. $\mathcal{L}_r$ and $\mathcal{L}_f$ indicate reconstruction and future prediction losses. TF denotes the transform operator: I (identity) and F (FFT). DQ denotes using dynamic queries with M (merge), R (remove), and S (split) operations. 
    }
    \label{tab:Abl_nuscenes}
    \centering
    \vspace{-8pt}
    \resizebox{\linewidth}{!}{
    \small
    \begin{tabular}{c|c|ccccc|ccc}
    \hline
      \quad  Epochs \quad & \quad \#Queries \quad & \quad $\mathcal{L}_r$ \quad & \quad $\mathcal{L}_f$ \quad  & \quad SSM \quad & \quad TF \quad & \quad DQ \quad& \quad  NDS$\uparrow$  \quad & \quad mAP$\uparrow$ \quad & \quad FPS$\uparrow$ \quad \\ \hline
     \multirow{2}{*}{24} & 900 & & & & &    & 55.4 & 44.9 & 14 \\
     & 400 & & & & &    & 55.0 & 44.3 & 24 \\
    \hline
     \multirow{5}{*}{24} & \multirow{5}{*}{900} & & & $\checkmark$ &  I &   & 55.5 & 45.0 & 11 \\
     & & & & $\checkmark$ &   F & & 55.6 & 45.1 & 10  \\ 
     & & $\checkmark$ &  & $\checkmark$ &    I &   &  55.8 & 45.7 & 11 \\ 
     & & $\checkmark$ & $\checkmark$ & $\checkmark$ &    I &   & 56.0 & 45.9 & 11 \\ 
     & & $\checkmark$ & $\checkmark$ & $\checkmark$ &    I+F &   & 56.1 & 46.1 & 10 \\\hline
      \multirow{3}{*}{24} & {900 $\rightarrow$ 700} & $\checkmark$ & $\checkmark$ &  $\checkmark$ & I & M  & 55.8 & 46.0 & 21 \\
      & {900 $\rightarrow$ 269} & $\checkmark$ & $\checkmark$ & $\checkmark$ &   I & M+R  & 55.9 & 45.9 & 35 \\
     & {900 $\rightarrow$ 269} & $\checkmark$ & $\checkmark$ & $\checkmark$ &   I & M+R+S  & 55.9 & 46.0 & 35 \\ 
    \hline
     24 & 900 $\rightarrow$ 269 & $\checkmark$ & $\checkmark$ & $\checkmark$ & I+F  &  M+R+S & \textbf{56.2}  & \textbf{46.2} & 33  \\ 
        \hline
    \end{tabular}
    }
    \vspace{-0pt}
\end{table*}

\subsection{Datasets and Metrics}
\noindent\textbf{nuScenes\cite{caesar2020nuscenes}.} This dataset serves as a benchmark for various autonomous driving tasks, including 3D object detection, tracking, and planning, etc. It features a comprehensive collection of large-scale, multi-modal data from 6 surround-view cameras, 1 Lidar, and 5 radars, totaling 1000 videos. The official nuScenes dataset is split into 700 videos for training, 150 for validation, and 150 for testing. An online platform provided by \cite{caesar2020nuscenes} allows for the evaluation of results on the test videos. Each video is approximately 20 seconds long, with key samples annotated every 0.5 seconds. The dataset includes 1.4 million annotated 3D bounding boxes across 10 classes for the 3D object detection task. 

\noindent\textbf{Metrics.} The standard evaluation metrics for 3D object detection task are mean Average Precision (mAP) and five true positive (TP) metrics: Average Translation Error (ATE), Average Scale Error (ASE), Average Orientation Error (AOE), Average Velocity Error (AVE), and Average Attribute Error (AAE). Overall performance is assessed using the nuScenes Detection Score (NDS), a composite of these metrics.



\subsection{Main Evaluation Results}
\subsubsection{nuScenes val split} In Table~\ref{tab:val_nuscenes}, we compare our proposed \ours and the current (SOTA) methods on the validation split of nuScenes. In this case, \ours dynamically reduces the number of queries from 900 to 269, with 269 being the minimum number of queries. Additionally, we utilize ResNet weights pretrained on nuImages~\cite{caesar2020nuscenes}.

Using ResNet50 as the backbone and an input size of 704$\times$256, \ours achieves 56.2 NDS and 46.2 mAP, outperforming the existing SOTA SparseBEV~\cite{liu2023sparsebev} by 0.4 NDS and 1.4 mAP. Moreover, \ours runs at 33 frames per second (FPS) on an Nvidia RTX-3080 GPU, more than 20\% faster than SparseBEV (27 FPS); see Fig.~\ref{fig:motivation} for more accuracy-efficiency comparisons. 
When a larger backbone of ResNet101 as well as a larger input size of 1408$\times$512 are used, \ours consistently outperforms existing SOTA methods.

\subsubsection{nuScenes test split} 
We report results on nuScenes official test benchmark in Table~\ref{tab:test_nuscenes}. With the commonly used V2-99~\cite{lee2020centermask} backbone pretrained with DD3D~\cite{park2021pseudo}, \ours achieves 65.1 NDS and 56.8 mAP, without future frames, test-time adaptation, ensembles, or any bells and whistles, outperforming latest methods such as RayDN and SparseBEV. Notably, \ours surpasses SparseBEV by a significant margin of 2.4 NDS and 2.8 mAP. It is important to note that SparseBEV~\cite{liu2023sparsebev} uses a fixed, large number of 1600 queries to achieve 62.7 NDS and 54.3 mAP. In contrast, \ours reduces the number of queries from 1600 to 729, thereby lowering computational complexity while still achieving 65.3 NDS and 57.4 mAP.

\subsection{Ablation Study and Further Evaluation}
We conduct ablation study on the nuScenes val split, starting with SparseBEV with ResNet50 backbone as the baseline. Table~\ref{tab:Abl_nuscenes} summarizes the results and showcases the effect of each component. In the table, 
SSM indicates the use of state-space modeling. $\mathcal{L}_r$ and $\mathcal{L}_f$ indicate the the use of losses in Eq.~(\ref{eq:mamba loss}). TF denotes the transform operator: I (identity) and F (FFT). DQ denotes using dynamic queries with M (merge), R (remove), and S (split) operations. 

\subsubsection{Number of queries} We train the baseline with 900 and 400 queries. When reducing the queries from 900 to 400, we observe a 0.56 drop in mAP and an increase in FPS from 14 to 24; see the first part of Table~\ref{tab:Abl_nuscenes}. While simply reducing the number of queries improves inference speed, it also degrades the detection performance of the baseline.

\subsubsection{State-space learning} Naively introducing an SSM like Mamba does not lead to significant improvement of detection performance. When we further introduce our proposed auxiliary training tasks of iterative future prediction and masked reconstruction, we see that the state-space learning becomes more effective, generating a substantial improvement of 0.73 in NDS and 1.25 in mAP as compared to the baseline. These results are summarized in the second part of Table~\ref{tab:Abl_nuscenes}.

\subsubsection{Dynamic queries} In the third part, we introduce merge, remove, and split one by one. When using all three operations and dynamically reducing the number of queries from 900 to 269, we achieve an NDS score almost identical to that of the baseline which uses 900 queries, yet the FPS increases from 11 to 35. With all the proposed designs, \ours achieves an NDS of 56.2, an mAP of 46.2, and an FPS of 33.

\subsubsection{Tracking} To further assess the temporal learning of \ours, we perform tracking and compare with SparseBEV~\cite{liu2023sparsebev} and StreamPETR~\cite{wang2023exploring} on the nuScenes validation split, and report the performance based on AMOTA (average multi-object tracking accuracy) and AMOTP (average multi-object tracking precision). We use ResNet50 as the backbone for all three models.

Table~\ref{tab:Abl_tracking} shows that \ours considerably outperforms SparseBEV and StreamPETR. This indicates that our proposed state-space learning effectively captures the spatial-temporal as well as motion information in the scene.

\begin{table}[h!]
    \caption{3D object tracking comparison on nuScenes val.}
	\label{tab:Abl_tracking}
    \centering
    \vspace{-6pt}
    \resizebox{\linewidth}{!}{
    \small
    \begin{tabular}{l|ccc}
    \hline
        Metric & StreamPETR\cite{wang2023exploring}  & SparseBEV\cite{liu2023sparsebev} &  \ours (ours) \\ \hline
         AMOTA$\uparrow$ & 33.7  & 33.9 &  35.1    \\ 
         AMOTP$\downarrow$  & 1.394 & 1.356  & 1.287  \\
        \hline
    \end{tabular}
    }
    \vspace{-0.0em}
\end{table}

\section{Conclusions}
In this paper, we introduce DySS, a novel and efficient method for 3D object detection. Our approach leverages state-space modeling to effectively capture the rich spatial-temporal information present in multi-camera video frames. By employing iterative future prediction and masked reconstruction, we train the state-space model (SSM) to better capture motion and correspondence cues. Additionally, we propose a dynamic query scheme that updates detection queries based on the learned state-space features.

{
    \small
    \bibliographystyle{ieeenat_fullname}
    \bibliography{main}
}


\end{document}